\documentclass[letterpaper, 10 pt, conference]{ieeeconf}  

\IEEEoverridecommandlockouts                              

\usepackage{graphics} 
\usepackage{epsfig} 
\usepackage{mathptmx} 
\usepackage{amsmath} 
\usepackage{amssymb}  
\usepackage{colortbl}
\usepackage{color}
\definecolor{tabcolor}{rgb}{1.0,0.0,0.0}
\usepackage{bm}
\usepackage{url}
\usepackage{threeparttable}
\usepackage{tabularx}
\usepackage{multirow}
\usepackage{booktabs}
\usepackage{flushend}
\title{\LARGE \bf
Optimization-Based Visual-Inertial SLAM \\ Tightly Coupled with Raw GNSS Measurements
}

\author{Jinxu Liu, Wei Gao* and Zhanyi Hu
\thanks{This work was supported in part by the National Key R\&D Program of China (2016YFB0502002), and in part by the Natural Science Foundation of China (61991423, 61872361). *Corresponding Author: Wei
Gao.}
\thanks{
All authors are with National Laboratory of Pattern Recognition, Institute of Automation, Chinese Academy of Sciences, and with School of Artificial Intelligence, University of Chinese Academy of Sciences, China. {\tt\small \{jinxu.liu,wgao,huzy\}@nlpr.ia.ac.cn}}}

\begin{document}

 \copyright 2021 IEEE. Personal use of this material is permitted.  Permission from IEEE must be obtained for all other uses, in any current or future media, including reprinting/republishing this material for advertising or promotional purposes, creating new collective works, for resale or redistribution to servers or lists, or reuse of any copyrighted component of this work in other works.

\maketitle
\thispagestyle{empty}
\pagestyle{empty}

\begin{abstract}

Unlike loose coupling approaches and the EKF-based approaches in the literature, we propose an optimization-based visual-inertial SLAM tightly coupled with raw Global Navigation Satellite System (GNSS) measurements, a first attempt of this kind in the literature to our knowledge. More specifically, reprojection error, IMU pre-integration error and raw GNSS measurement error are jointly minimized within a sliding window, in which the asynchronism between images and raw GNSS measurements is accounted for. In addition, issues such as marginalization, noisy measurements removal, as well as tackling vulnerable situations are also addressed. Experimental results on public dataset in complex urban scenes show that our proposed approach outperforms state-of-the-art visual-inertial SLAM, GNSS single point positioning, as well as a loose coupling approach, including scenes mainly containing low-rise buildings and those containing urban canyons. 

\end{abstract}

\section{INTRODUCTION}

Positioning in outdoor scenes has long been a concerned task in the realms such as autonomous driving. Global Navigation Satellite System (GNSS) has been widely exploited for decades, due to its ability to acquire geodetic coordinates and its stable performance in open area. However, in \emph{urban canyons} where there are lots of skyscrapers in the surroundings, GNSS positioning results dramatically deteriorate. On the other hand, V-SLAM or VI-SLAM approaches do not suffer from \emph{urban canyons}, and their local accuracy is much higher than that of single point GNSS positioning.

However, V-SLAM and VI-SLAM also have their inherent shortcomings. Besides the deterioration of accuracy in low-textured scenes and the scenes with many moving objects, they typically suffer from drift. And more importantly, since global position is unobservable for SLAM approaches, what can be obtained from SLAM approaches are the poses in a local world frame instead of a geodetic coordinate frame. Hence GNSS positioning and V-SLAM or VI-SLAM approaches are complementary, due to their different global and local accuracy, as well as their different vulnerable scenes. Hence fusing vision, IMU and GNSS is promising for accurate global positioning in complex urban scenes.
\section{RELATED WORK}

\subsection{Loose Coupling Approaches}

Loose coupling approaches firstly derive poses from feature points, or the positions and velocities from raw GNSS measurements, or both of the above two quantities. Then the derived quantities rather than raw measurements are involved in the fusion stage. Some approaches \cite{schreiber2016vehicle}, \cite{gakne2018tightly} derive poses from feature points, and fuse them with raw GNSS measurements such as pseudorange and Doppler shift. Conversely, some approaches \cite{yu2019gps}, \cite{li2020intermittent}, \cite{cioffi2020tightly} derive positions from raw GNSS measurements, and fuse them with feature points from the images. Other approaches \cite{rehder2012global}, \cite{mascaro2018gomsf}, \cite{yan2018image}, \cite{qin2019b}  apply loose coupling on both the vision side and the GNSS side. Some approaches \cite{gakne2018tightly}, \cite{wen2020tightly} use visual information to aid GNSS positioning by judging which satellites are visible. The above loose coupling approaches are not optimal in the sense that the available information is not fully exploited. 

\subsection{Tight Coupling Approaches}

This category of approaches employs tight coupling on both the vision side and the GNSS side, i.e., both the feature points from the image and the raw GNSS measurements serve as measurements in either EKF-based or optimization-based framework. The EKF-based approaches in \cite{won2014gnss}, \cite{won2014selective} tightly couple vision, IMU and single point GNSS. They perform state prediction using the IMU measurements, and update the states using the feature points as well as the pseudorange and Doppler shift measurements. Similar EKF-based framework is adopted in \cite{li2019tight}, while \cite{li2019tight} utilizes double-differenced GNSS instead of single point GNSS. 
Although double-differenced GNSS provides measurements for more accurate positioning, it requires an additional base station in comparison with single point GNSS.

Besides EKF-based approaches, optimization-based approaches are emerging as well. Compared with EKF, batch optimization allows for the reduction of error through relinearization in visual-inertial navigation systems \cite{huang2019visual}  and proves to have better performance in GNSS-IMU fusion task \cite{wen2020time}. Therefore, optimization is a promising solution to fuse vision, IMU and GNSS measurements. The optimization-based approach \cite{gong2019tightly} tightly couples feature points with pseudorange from single point GNSS in bundle adjustment. However, Doppler shift is not uitilized, and IMU measurements are not tightly coupled in \cite{gong2019tightly}, and only the magnetometer in IMU is exploited, to determine the direction of the local world frame. Nevertheless, tightly coupling IMU measurements can eliminate the effect of the asynchronism between vision and GNSS measurements, thus the low-speed motion assumption applied in \cite{gong2019tightly} is not needed in this case.

In this paper we propose an optimization-based approach that tightly couples vision, IMU and single point raw GNSS measurements including pseudoranges and Doppler shift, a first attempt in literature to our knowledge.\footnote{For more details of this paper, the reader may see our arXiv technical report \url{https://arxiv.org/abs/2010.11675v4}.}

\section{FRAMES AND NOTATIONS}

\begin{figure}[htb]
\centering
\includegraphics[width=6.5cm]{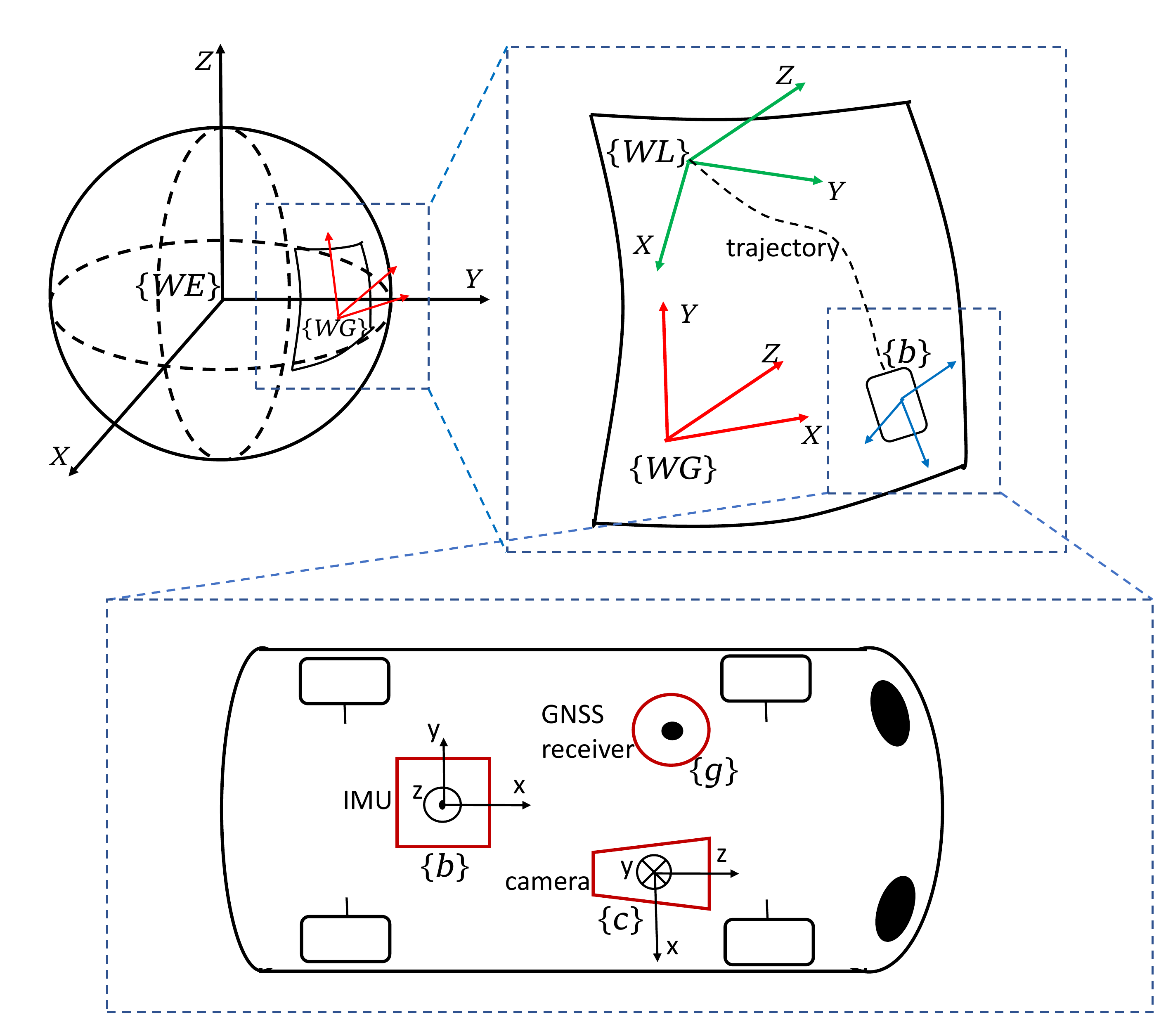}
\caption{Illustration of world frames and sensor frames. The globe in the upper left of the figure is the earth globe. The origin of the ECEF frame $\{WE\}$ is on the center of the earth. The X axis of the ECEF frame points to the point of intersection between the prime meridian and equator, and the Z axis points to the North Pole. The ground ENU frame $\{WG\}$ and the local world frame for SLAM $\{WL\}$ both locate on the ground, with their Z axes pointing upward.}
\label{fig_frames}
\end{figure}

In this paper, the world frames include the earth-centered, earth-fixed (ECEF) frame, the ground east-north-up (ENU) frame, and the local world frame for SLAM. The sensor frames include the camera frame, the IMU frame and the GNSS receiver frame. The above world frames and sensor frames are illustrated in Fig.\ref{fig_frames}. We use $(\cdot)^{WE}$, $(\cdot)^{WG}$ and $(\cdot)^{WL}$ to denote the ECEF frame, the ground ENU frame and the local world frame for SLAM respectively, and use $(\cdot)^c$, $(\cdot)^b$ and $(\cdot)^g$ to denote the camera frame, IMU frame and GNSS receiver frame respectively. Specifically, we use $(\cdot)^{b_k}$ and $(\cdot)^{g_k}$ to denote the IMU frame and GNSS receiver frame corresponding to the $k^{th}$ image, and slightly abuse the symbols using $(\cdot)^{b_t}$ and $(\cdot)^{g_t}$ to denote the IMU frame and GNSS receiver frame at some moment $t$.

Let $\mathbf{R}_A^B$ denote the rotation matrix that takes a vector in frame $\{A\}$ to frame $\{B\}$, and $\mathbf{q}_A^B$ is its quaternion form. $\mathbf{p}_A^B$ is the coordinate of the origin point of frame $\{A\}$ in frame $\{B\}$, and $\mathbf{v}_A^B$ is the velocity of the origin point of frame $\{A\}$ measured in frame $\{B\}$. Note that for GNSS receiver frame, only the coordinate of its origin point in frame $\{WL\}$ matters in our approach, and the orientations of its axes can be regarded as arbitrary. Let $\mathbf{b}_{a_k}$ and $\mathbf{b}_{\omega_k}$ denote the accelerometer bias and gyroscope bias corresponding to image $k$ respectively. Moreover, we use $[\cdot]_{\times}$ to denote the skew symmetric matrix corresponding to a vector.

\section{OUR APPROACH}
\subsection{GNSS-SLAM Initialization}
\label{subsec_initializaiton}
After our system starts, the initialization for VI-SLAM is performed, and the nonlinear optimization for VI-SLAM starts performing temporarily following the routine of \cite{qin2018vins}. Meanwhile, the GNSS single point positioning is performed. Hence we have obtained the VI-SLAM trajectory and the GNSS single point positioning trajectory respectively. At GNSS-SLAM initialization stage we align the above two trajectories, in order to compute the value of $\mathbf{R}^{WE}_{WG}$ and $\mathbf{p}^{WE}_{WG}$, as well as the initial value of $\mathbf{R}^{WG}_{WL}$ and $\mathbf{p}^{WG}_{WL}$.

Firstly, we select one position calculated by GNSS single point positioning as the reference point $\mathbf{p}^{WE}_{ref}$, which serves as the origin point of the ground ENU frame. In our implementation, the third GNSS single point positioning result since the system starts is selected as the reference point. Given the reference point, $\mathbf{R}^{WE}_{WG}$ and $\mathbf{p}^{WE}_{WG}$ can be computed straightforwardly \cite{xie2009principles}. Then the GNSS single point positioning trajectory is converted from the ECEF frame to the ground ENU frame as
\begin{equation}
\label{eqn8}
\small
\mathbf{p}^{WG}_g = {\mathbf{R}^{WE}_{WG}}^T(\mathbf{p}^{WE}_g - \mathbf{p}^{WE}_{WG}).
\end{equation}
Hence we have two sets of positions $\{\mathbf{p}^{WG}_{g_l}|l=0\dots L-1\}$ and $\{\mathbf{p}^{WL}_{g_l}|l=0\dots L-1\}$, of which the former includes the positions from GNSS single point positioning in frame $\{WG\}$, and the latter includes the positions estimated from VI-SLAM in frame $\{WL\}$. Note that the positions from VI-SLAM are in fact $\mathbf{p}^{WL}_{b_l}$, but at the initialization stage we neglect the translation from IMU frame to GNSS receiver frame, thus we regard $\mathbf{p}^{WL}_{b_l}$ as $\mathbf{p}^{WL}_{g_l}$. At this stage interpolation is performed to align the timestamps of the above two sets of positions. At last a 5-Degrees of Freedom(DoF) alignment between the above two sets is performed to compute the initial value of $\mathbf{R}^{WG}_{WL}$ and $\mathbf{p}^{WG}_{WL}$, by minimizing the following cost:
\begin{equation}
\label{eqn9}
\small
\min_{s, \mathbf{R}^{WG}_{WL}, \mathbf{p}^{WG}_{WL}}\sum_{l=0}^{L-1}{\parallel \mathbf{p}^{WG}_{g_l} - s\mathbf{R}^{WG}_{WL}\mathbf{p}^{WL}_{g_l} -  \mathbf{p}^{WG}_{WL}\parallel^2}.
\end{equation}
The scale parameter $s$ is estimated because the VI-SLAM is sensitive to scale drift in degenerate cases \cite{wu2017vins}. Note that $\mathbf{R}^{WG}_{WL}$ only contains one DoF, because the Z axes of both frame $\{WG\}$ and frame $\{WL\}$ point upward. 

\subsection{Tightly Coupled Optimization}
\label{ssec:optimization}
After GNSS-SLAM initialization, raw GNSS measurements, i.e. pseudoranges and Doppler shift, are integrated into the optimization. The states to be estimated include
\begin{equation}
\label{eqn10}
\small
\begin{split}
\mathcal{X} = [\mathbf{x}_0\dots\mathbf{x}_{K-1},f_0\dots f_{M-1},\mathbf{T}^b_c, \mathbf{T}^{WG}_{WL}, \bm{\delta t}_0^r...\bm{\delta t}_{E-1}^r,\dot{\bm{\delta t}}_0^r...\dot{\bm{\delta t}}_{E-1}^r],
\end{split}
\end{equation}
where $K$ is the number of images in the sliding window, $M$ is the number of landmarks, and $E$ is the number of GNSS epochs in the sliding window. $f$ is the inverse depth of one landmark in camera frame, and
\begin{equation}
\label{eqn11}
\small
\begin{split}
\mathbf{x}_k&=\begin{bmatrix} \mathbf{p}^{WL}_{b_k},\mathbf{v}^{WL}_{b_k},\mathbf{q}^{WL}_{b_k},\mathbf{b}_{a_k},\mathbf{b}_{\omega_k} \end{bmatrix},k=0\dots K-1,\\
\mathbf{T}^b_c&=\begin{bmatrix}\mathbf{R}^b_c,\mathbf{p}^b_c\end{bmatrix}, \mathbf{T}^{WG}_{WL}=\begin{bmatrix}\mathbf{R}^{WG}_{WL},\mathbf{p}^{WG}_{WL}\end{bmatrix},\\
\bm{\delta t}_e^r &= \begin{bmatrix}\{\delta t^r_{j,e}\vert \forall j \in \mathcal{O}(e)\} \end{bmatrix},e=0\dots E-1,\\ \dot{\bm{\delta t}}_e^r &= \begin{bmatrix}\{\dot{\delta t}^r_{j,e}\vert \forall j \in \mathcal{O}(e)\} \end{bmatrix},e=0\dots E-1,
\end{split}
\end{equation}
where $\mathcal{O}(e)$ is the set of observable GNSS constellations such as GPS, GLONASS, etc. at epoch $e$, $\delta t^r_{j,e}$ and $\dot{\delta t}^r_{j,e}$ are the receiver clock bias and clock drift w.r.t GNSS constellation $j$ at epoch $e$ respectively, and $\bm{\delta t}^r_e$ and $\dot{\bm{\delta t}}^r_e$ are the vectors containing the receiver clock biases and clock drifts for all the observable constellations at epoch $e$ respectively. Compared with VINS-mono \cite{qin2018vins}, the introduced additional parameters include the transformation from the local world frame to ground ENU frame, as well as clock biases and clock drifts. The transformation from ground ENU frame to ECEF frame $(\mathbf{R}^{WE}_{WG}, \mathbf{p}^{WE}_{WG})$ is treated as a known quantity rather than a parameter, because the ground ENU frame can be located anywhere on the ground that is not too far away from the origin of the local world frame for SLAM, for example 1 km. 
Although GNSS measurements and images are collected at different moments, we do not introduce additional parameters concerning pose and velocity at GNSS measurement moments. In other words, we incorporate intermediate GNSS measurements, which is inspired by \cite{chng2019outlier}. Note that just like in the GNSS-SLAM initialization stage, $\mathbf{R}^{WG}_{WL}$ only has 1 DoF, so it is parameterized only by the rotation angle around the Z axis, which writes as $\kappa^{WG}_{WL}$.

The cost function $c(\mathcal{X})$ to be minimized is
\begin{equation}
\label{eqn12}
\small
\begin{split}
c(\mathcal{X})&=c^{reproj}(\mathcal{X})+c^{IMU}(\mathcal{X})+\sum_{e=0}^{E-1}\sum_{i\in\mathcal{S}(e)} {\mathbf{e}^P_{i,e}}^T\mathbf{W}^P\mathbf{e}^P_{i,e}\\
&+\sum_{e=0}^{E-1}\sum_{i\in\mathcal{S}(e)} {\mathbf{e}^D_{i,e}}^T\mathbf{W}^D\mathbf{e}^D_{i,e}+{\mathbf{e}^{marg}}^T\mathbf{e}^{marg},
\end{split}
\end{equation}
where $E$ is the number of GNSS epochs in the sliding window and $\mathcal{S}(e)$ is the set of observable GNSS satellites at epoch $e$. $\mathbf{e}^P_{i,e}$ and $\mathbf{e}^D_{i,e}$ are the pseudorange residual and the Doppler shift residual for satellite $i$ observed at epoch $e$ respectively. $\mathbf{e}^{marg}$ denotes the marginalization residual. The reprojection factors $c^{reproj}(\mathcal{X})$ and IMU factors $c^{IMU}(\mathcal{X})$ are identical to those in VINS-mono \cite{qin2018vins}, and the readers may refer to (14), (16) and (17) in VINS-mono \cite{qin2018vins} for details. The IMU pre-integration performed in our proposed approach is also identical to that in VINS-mono \cite{qin2018vins}.

As for the GNSS factors, the pseudorange residual $\mathbf{e}^P_{i,e}$ and the Doppler shift residual $\mathbf{e}^D_{i,e}$ write as
\begin{equation}
\label{eqn13}
\small
\begin{split}
\mathbf{e}^P_{i,e}&=\parallel \mathbf{p}^{WE}_{s_i,e}-(\mathbf{R}^{WE}_{WG}\mathbf{R}^{WG}_{WL}\mathbf{p}^{WL}_{g_{m(e)}}+\mathbf{R}^{WE}_{WG}\mathbf{p}^{WG}_{WL}+\mathbf{p}^{WE}_{WG})\parallel +c\delta t_{j(i),e}^r\\ &-c\delta t_{i,e}^s-\rho_{i,e},\\
\mathbf{e}^D_{i,e}&= \mathbf{l}_{i,e}^T(\mathbf{v}^{WE}_{s_i,e}-\mathbf{R}^{WE}_{WG}\mathbf{R}^{WG}_{WL}\mathbf{v}^{WL}_{g_{m(e)}})+c\dot{\delta t}_{j(i),e}^r-c\dot{\delta t}_{i,e}^s-(-\lambda_{i,e}D_{i,e}),
\end{split}
\end{equation}
where 
\begin{equation}
\label{eqn14}
\small
\begin{split}
\mathbf{l}_{i,e}&=\frac{\mathbf{p}^{WE}_{s_i,e}-(\mathbf{R}^{WE}_{WG}\mathbf{R}^{WG}_{WL}\mathbf{p}^{WL}_{g_{m(e)}}+\mathbf{R}^{WE}_{WG}\mathbf{p}^{WG}_{WL}+\mathbf{p}^{WE}_{WG})}{\parallel \mathbf{p}^{WE}_{s_i,e}-(\mathbf{R}^{WE}_{WG}\mathbf{R}^{WG}_{WL}\mathbf{p}^{WL}_{g_{m(e)}}+\mathbf{R}^{WE}_{WG}\mathbf{p}^{WG}_{WL}+\mathbf{p}^{WE}_{WG})\parallel},\\
\mathbf{p}^{WL}_{g_{m(e)}}&=\mathbf{p}^{WL}_{b_k}+\mathbf{v}^{WL}_{b_k}\Delta t_{k,m(e)}+\mathbf{R}^{WL}_{b_k}(\bm{\alpha}^{b_k}_{b_{m(e)}}+\mathbf{R}(\bm{\gamma}^{b_k}_{b_{m(e)}})\mathbf{p}^b_g)\\
&-\frac12\mathbf{g}^{WL}\Delta t_{k,m(e)}^2,\\
\mathbf{v}^{WL}_{g_{m(e)}}&=\mathbf{v}^{WL}_{b_k}+\mathbf{R}^{WL}_{b_k}(\bm{\beta}^{b_k}_{b_{m(e)}}-\mathbf{R}(\bm{\gamma}^{b_k}_{b_{m(e)}})[\mathbf{p}^b_g]_{\times}(
\hat{\bm{\omega}}_{m(e)}-\mathbf{b}_{\omega_k}))\\
&-\mathbf{g}^{WL}\Delta t_{k,m(e)}.
\end{split}
\end{equation}
In (\ref{eqn13}) and (\ref{eqn14}), $c$ denotes the speed of light, $\mathbf{p}^{WE}_{s_i,e}$ and $\delta t_{i,e}^s$ are the position and clock bias of satellite $i$ at GNSS epoch $e$ respectively, and $\mathbf{v}^{WE}_{s_i,e}$ and $\dot{\delta t}_{i,e}^s$ are the velocity and clock drift of satellite $i$ at epoch $e$ respectively. The above four quantities are computed directly from the broadcast emphemeris. The subscript $j(i)$ means the constellation corresponding to satellite $i$, and the subscript $m(e)$ denotes the measurement moment of GNSS epoch $e$. 
The parameters $\delta t^r_{j(i),e}$ and $\dot{\delta t}^r_{j(i),e}$ are the receiver clock bias and clock drift w.r.t GNSS constellation $j(i)$ at epoch $e$ respectively. $\rho_{i,e}$, $\lambda_{i,e}$ and $D_{i,e}$ are the pseudorange measurement, wavelength of the carrier, and Doppler shift measurement of satellite $i$ at epoch $e$ respectively. 
$\Delta t_{k,m(e)}$ is the time interval from the exposure moment of image $k$ to the moment $m(e)$. $\mathbf{p}^b_g$ is the translational component of IMU-GNSS receiver extrinsic parameters, which is assumed to have been accurately calibrated beforehand in this paper. $\bm{\alpha}^{b_k}_{b_{m(e)}}$, $\bm{\beta}^{b_k}_{b_{m(e)}}$ and $\bm{\gamma}^{b_k}_{b_{m(e)}}$ are the nominal states coming from IMU pre-integration from the exposure moment of image $k$ to the moment $m(e)$, and they correspond to displacement, change in velocity, and relative rotation respectively. $\mathbf{R}(\bm{\gamma}^{b_k}_{b_{m(e)}})$ is the rotation matrix converted from the rotation quaternion $\bm{\gamma}^{b_k}_{b_{m(e)}}$. For the details of IMU-preintegration the readers may refer to (3) and (5) in VINS-mono \cite{qin2018vins}. $\hat{\bm{\omega}}_{m(e)}$ is the angular velocity measurement from IMU at moment $m(e)$, and $\mathbf{g}^{WL}$ is the gravity vector in frame $\{WL\}$, which writes as $\begin{bmatrix} 0&0&g\end{bmatrix}^T$ with $g$ being the magnitude of gravity. In (\ref{eqn14}), an approximation is applied that the gyroscope bias at image exposure moment $\mathbf{b}_{w_k}$ substitutes for the gyroscope bias at GNSS measurement moment $m(e)$, because gyroscope bias is a slow time-varying quantity.

In fact, $\mathbf{W}^P$ and $\mathbf{W}^D$ in (\ref{eqn12}) are both $1\times 1$ matrices which can be regarded as scalars. In our implementation, we adopt $\mathbf{W}^P=1$ and $\mathbf{W}^D=4$, assuming the median errors of pseudorange measurements and pseudorange rates converted from Doppler shift measurements to be $1m$ and $0.5m/s$ respectively. The optimization is performed in a sliding window by minimizing the cost function $c(\mathcal{X})$ using Ceres Solver \cite{ceres-solver}.

\subsection{Sliding Window and Marginalization}
\begin{figure}[htb]
\centering
\includegraphics[width=8.0cm]{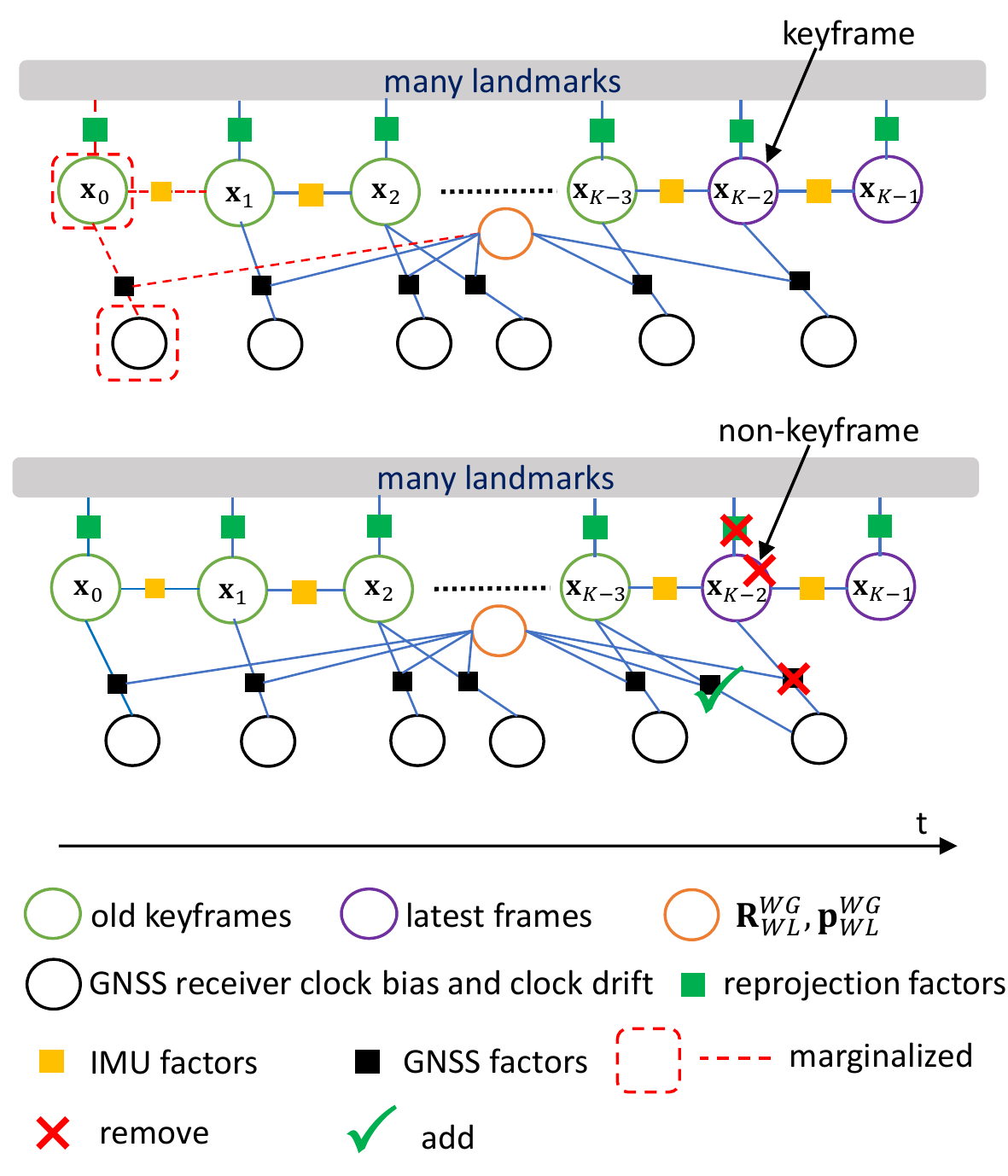}
\caption{Marginalization in sliding window. The hollow circles represent the parameters (nodes), and the solid rectangles represent the factors (edges). The camera-IMU extrinsic parameters and the marginalization factor are not shown here for clarity of explanation.}
\label{fig_marginalization}
\end{figure}

In the sliding window in our approach illustrated in Fig.\ref{fig_marginalization}, the parameters related to GNSS, i.e. GNSS receiver clock bias and clock drift, are \emph{attached} to the image frame just before it. There may exist more than one GNSS parameter node attached to a certain image frame.
If the second latest frame is a keyframe, we keep it in the sliding window, and marginalize the oldest keyframe and the GNSS parameters and feature points attached to it, as well as the reprojection factors, IMU factors and GNSS factors related to them. If the second latest frame is not a keyframe, we simply remove the frame and all its corresponding visual measurements. However, the GNSS parameters attached to the second latest frame will not be removed, and they will be attached to the third latest frame instead, i.e. removed and added in Fig.\ref{fig_marginalization}. The GNSS factors that are originally attached to the second latest frame will be changed, for the reason that the IMU pre-integration in the above GNSS factors need to be recomputed because their starting moments will switch from the second latest frame to the third latest frame. The GNSS measurements in the above GNSS factors remain unchanged.

The marginalization in our approach is carried out using Schur-Complement. The readers may refer to \cite{leutenegger2015keyframe} and \cite{qin2018vins} for details of applying marginalization. Note that since marginalization is applied, the GNSS measurements that have slid out of the sliding window still take effect on the state estimation in the current sliding window. Therefore $\mathbf{R}^{WG}_{WL}$ and $\mathbf{p}^{WG}_{WL}$ are still observable even if there are no GNSS measurements in the current sliding window. 

\subsection{Removing Noisy GNSS Measurements}
\label{subsec_better}
\begin{figure}[htb]
\begin{minipage}[b]{.48\linewidth}
  \centering
  \centerline{\includegraphics[width=4.4cm]{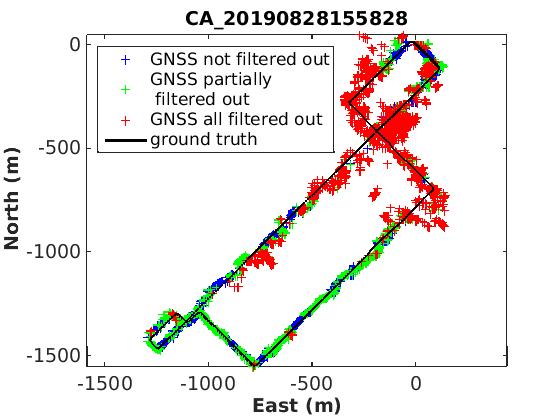}}
  \centerline{(a) east and north}\medskip
\end{minipage}
\hfill
\begin{minipage}[b]{0.48\linewidth}
  \centering
  \centerline{\includegraphics[width=4.4cm]{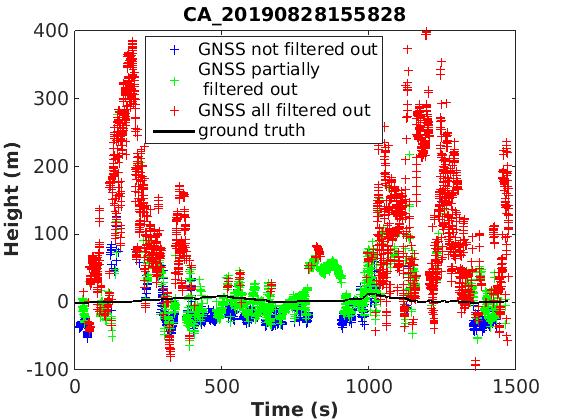}}
  \centerline{(b) height}\medskip
\end{minipage}

\caption{GNSS single point positioning results w.r.t east and north directions (a), and w.r.t height direction (b). The red, green and blue markers are epochs when GNSS measurements are all filtered out, partially filtered out and not filtered out respectively.}
\label{fig_quality}
\end{figure}

In urban canyons, raw GNSS measurement can be pretty noisy. Therefore, before the GNSS measurements enter into the sliding window for optimization, a filtering is carried out by one of the following two possible methods. The first method is to perform single point positioning, as well as velocity determination using raw Doppler shift measurements \cite{xie2009principles}. For the second method, if all GNSS measurements are excluded during the past 5 seconds, single point positioning and velocity determination using raw Doppler shift measurements are performed. Otherwise a joint optimization is performed inside a sliding window similar to Sect. \ref{ssec:optimization}, but the optimization here differs with Sect. \ref{ssec:optimization} in that only the GNSS measurements at the current epoch are involved in the optimization. For both methods, we compute the residuals of pseudoranges and Doppler shift measurements after the optimization. The pseudoranges and Doppler shifts are filtered out if their corresponding residuals exceed the threshold $T^P$ and $T^D$ respectively. If too few GNSS measurements are left after filtering out noisy measurements at a certain GNSS epoch, we will remove all the GNSS measurements at the epoch altogether. The first method has the advantage in speed and the second one has the advantage in accuracy. The reason we perform a GNSS single point positioning if all GNSS measurements are excluded during the past 5 seconds for the second method is that, in this way we prevent good GNSS measurements from being excluded because of the incorrectly estimated current position. A discussion on the above two methods is presented in Sect. \ref{ssec:discussion}. In our implementation, $T^P=10m$ and $T^D=3m/s$. Fig.\ref{fig_quality} shows the GNSS single point positioning results and the results of our filtering operation by the first method. From Fig.\ref{fig_quality} we can see that the GNSS epochs when measurements are all filtered out generally experience less accurate single point positioning results than the epochs when measurements are partially filtered out or not filtered out.

\subsection{Some Implementation Details}
\label{ssec:details}

\begin{table*}[htb]
\caption{Comparison of mean absolute error (MAE), root mean square error (RMSE) and trajectory completeness (TC) of resulting trajectories from different approaches}
\label{table_average_error}
\centering
\begin{threeparttable}
\begin{tabular}{*{11}{c}}

\bottomrule
\multirow{2}{*}{Data Sequence}&\multirow{2}{*}{Length}&\multirow{2}{*}{Approach}&\multicolumn{3}{c}{MAE in Translation (m)}&\multicolumn{3}{c}{MAE in Rotation (deg)}&RMSE in&\multirow{2}{*}{TC}\\
&&&X&Y&Z&Z&Y&X&Trans. (m)&\\
\midrule
\multirow{4}{*}{CA\_20190828155828}&\multirow{4}{*}{5.9km}&VINS-mono \cite{qin2018vins}&93.75&77.38&11.16&2.106&1.036&2.501&142.74&99.97\%\\
&&GNSS SPP&\textbf{6.294}&\textbf{5.269}&11.96&*&*&*&28.12&46.85\%\\
&&VINS-Fusion (with GNSS) \cite{qin2019b}&11.59&9.256&33.94&5.225&10.72&11.82&55.15&98.73\%\\
&&Proposed&7.063&7.637&\textbf{4.047}&\textbf{1.499}&\textbf{0.323}&\textbf{2.143}&\textbf{14.33}&99.97\%\\
\midrule
\multirow{4}{*}{CA\_20190828173350}&\multirow{4}{*}{3.2km}&VINS-mono \cite{qin2018vins}&49.17&30.79&\textbf{2.686}&2.251&\textbf{0.644}&2.206&64.84&99.76\%\\
&&GNSS SPP&88.00&48.29&181.26&*&*&*&861.72&50.73\%\\
&&VINS-Fusion (with GNSS) \cite{qin2019b}&15.69&15.80&30.17&21.39&26.59&16.87&51.59&99.77\%\\
&&Proposed&\textbf{2.897}&\textbf{11.48}&3.523&\textbf{1.006}&0.668&\textbf{2.098}&\textbf{14.34}&99.76\%\\
\midrule
\multirow{4}{*}{CA\_20190828184706}&\multirow{4}{*}{1.8km}&VINS-mono \cite{qin2018vins}&6.669&6.080&1.347&\textbf{2.608}&0.455&2.059&11.29&98.39\%\\
&&GNSS SPP&2.270&2.190&5.514&*&*&*&8.394&93.03\%\\
&&VINS-Fusion (with GNSS) \cite{qin2019b}&2.120&2.729&4.519&6.130&15.17&9.207&7.518&100.00\%\\
&&Proposed&\textbf{2.099}&\textbf{1.692}&\textbf{0.623}&4.671&\textbf{0.367}&\textbf{1.863}&\textbf{3.590}&98.39\%\\
\midrule
\multirow{4}{*}{CA\_20190828190411}&\multirow{4}{*}{1.0km}&VINS-mono \cite{qin2018vins}&7.128&7.816&\textbf{1.156}&2.191&1.266&2.499&11.88&99.49\%\\
&&GNSS SPP&3.100&3.117&11.03&*&*&*&15.72&100.00\%\\
&&VINS-Fusion (with GNSS) \cite{qin2019b}&6.698&2.833&10.05&10.61&10.08&8.418&18.96&99.37\%\\
&&Proposed&\textbf{1.604}&\textbf{1.476}&2.735&\textbf{2.135}&\textbf{0.814}&\textbf{2.151}&\textbf{4.455}&99.56\%\\
\bottomrule
\end{tabular}
\begin{tablenotes}
        \fontsize{8pt}{10pt}
		\item Here \emph{GNSS SPP} means GNSS single point positioning. \emph{Proposed} means the proposed approach in this paper. The symbol * means the result is not available. GNSS single point positioning only provides positioning results, and hence its rotational accuracy is not available.
\end{tablenotes}
\end{threeparttable}
\end{table*}
\begin{itemize}
\item[$\bullet$] At the initial stage just after GNSS-SLAM initialization, a prior factor constraining the parameters $(\mathbf{R}^{WG}_{WL},\mathbf{p}^{WG}_{WL})$ is added to the sliding window, because at the initial stage the marginalization factor does not contain enough information from GNSS measurements in the past, and thus the estimation of $(\mathbf{R}^{WG}_{WL},\mathbf{p}^{WG}_{WL})$ is sensitive to noises in GNSS measurements. After the GNSS factors in more than $T^N$ GNSS epochs have been marginalized, the prior factor is permanently removed from the sliding window. In our implementation, $T^N=30$.
\item[$\bullet$] According to \cite{wu2017vins} and \cite{liu2019visual}, VI-SLAM system is not well-constrained before the platform has made a turn for the first time. Therefore, we employ the strategy proposed in \cite{liu2019visual} but in a much simpler way that we start to adjust the camera-IMU extrinsic parameters $(\mathbf{R}^b_c,\mathbf{p}^b_c)$ in optimization after the platform has made a turn. Here a \emph{turn} means a motion with large rotation, e.g. the vehicle turning left or turning right at a crossroad.
\item[$\bullet$] It is a vulnerable scene for our approach when the car stops at a crossroad. In our implementation when the estimated speed of the second latest frame is under $0.5m/s$, the GNSS factors in the sliding window are temporarily inactivated, but they still participate in marginalization as if they were in the sliding window.
\end{itemize}
\section{EXPERIMENTS}
The experiments are conducted on the public dataset UrbanLoco \cite{wen2020urbanloco}, which is collected in highly urbanized areas in San Francisco and Hong Kong. The images in Hong Kong data are collected using a fisheye sky camera, and the images in San Francisco data are collected using six 360-degree view cameras. According to \cite{wen2020urbanloco}, the ground truth in UrbanLoco dataset is provided by Novatel SPAN-CPT, which is an RTK-IMU navigation system. According to \cite{wen2020urbanloco}, when the satellite observability is satisfactory, the ground truth error is within 2 cm, and the calibration certificate dictates that the error after 10 seconds of GNSS backout is within 12 cm. However, there does exist something wrong with the ground truth in the sequence CA\_20190828190411, and hence we have manually excluded the incorrect part of ground truth in the sequence CA\_20190828190411 according to the Google Earth. Our approach is evaluated on 4 sequences collected in San Francisco, because the other 3 sequences in San Francisco contain tunnels, which are very tough scenerios for both camera and GNSS signals, and VINS-mono \cite{qin2018vins} evaluation results on only the above 4 sequences are reported in the paper \cite{wen2020urbanloco}. The 4 sequences contain both urban canyons and scenes with low-rise buildings. All the
experiments presented are performed on a PC with Intel Core
i7 3.6GHz$\times$6 core CPU and 64GB memory.

\begin{figure}[htb]
\begin{minipage}[b]{.48\linewidth}
  \centering
  \centerline{\includegraphics[width=4.4cm]{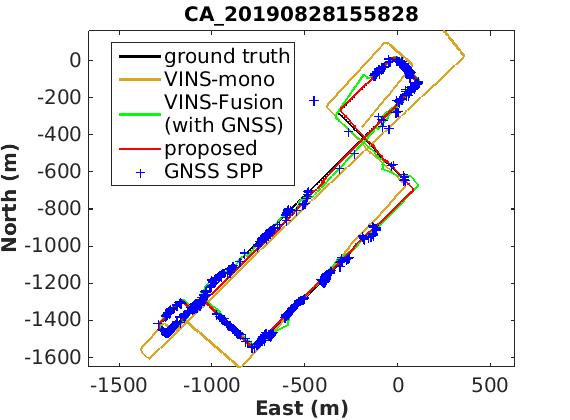}}
  \centerline{(a) CA\_20190828155828}\medskip
\end{minipage}
\hfill
\begin{minipage}[b]{0.48\linewidth}
  \centering
  \centerline{\includegraphics[width=4.4cm]{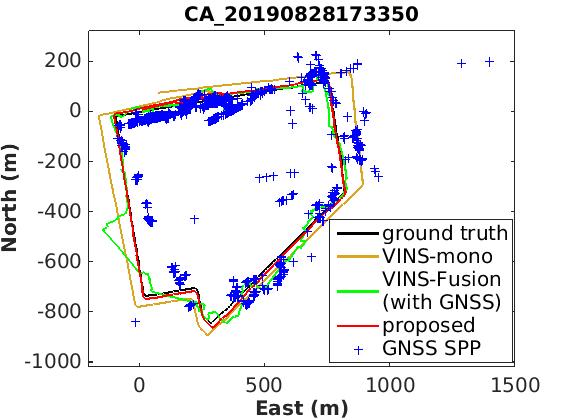}}
  \centerline{(b) CA\_20190828173350}\medskip
\end{minipage}
\begin{minipage}[b]{.48\linewidth}
  \centering
  \centerline{\includegraphics[width=4.4cm]{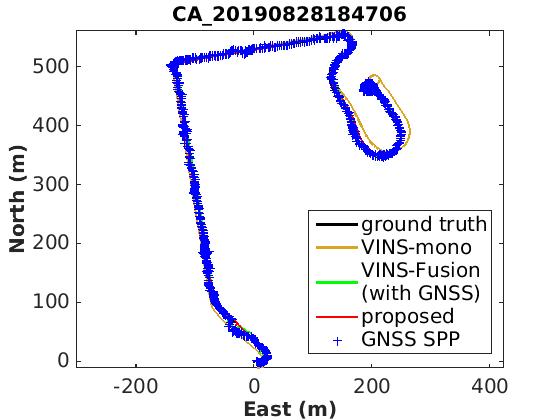}}
  \centerline{(c) CA\_20190828184706}\medskip
\end{minipage}
\hfill
\begin{minipage}[b]{0.48\linewidth}
  \centering
  \centerline{\includegraphics[width=4.4cm]{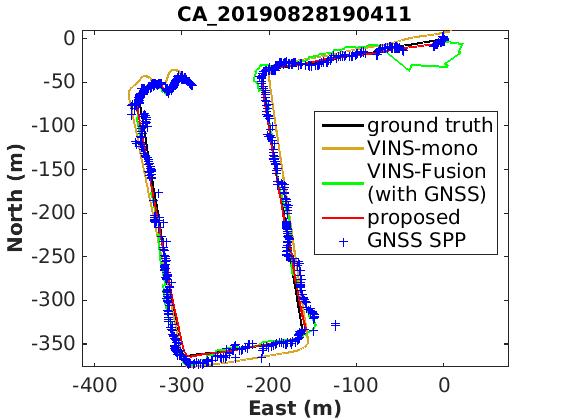}}
  \centerline{(d) CA\_20190828190411}\medskip
\end{minipage}
\caption{Comparison of trajectories in east and north directions. In figure (b) a few very noisy GNSS SPP results that lie outside the figure are omitted. All trajectories are aligned with the ground truth.}
\label{fig_tra}
\end{figure}

\begin{figure}[htb]
\begin{minipage}[b]{.48\linewidth}
  \centering
  \centerline{\includegraphics[width=4.4cm]{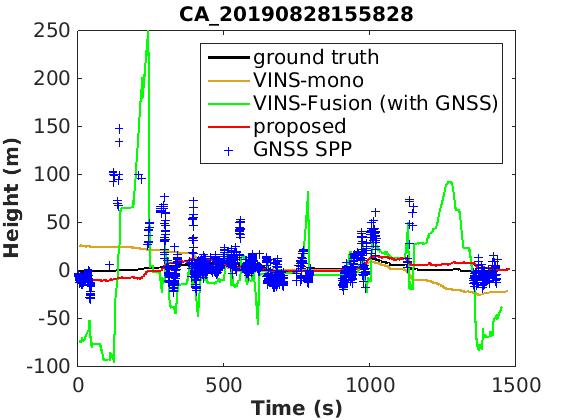}}
  \centerline{(a) CA\_20190828155828}\medskip
\end{minipage}
\hfill
\begin{minipage}[b]{0.48\linewidth}
  \centering
  \centerline{\includegraphics[width=4.4cm]{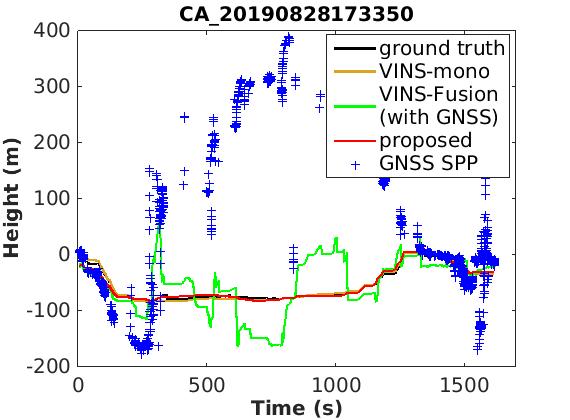}}
  \centerline{(b) CA\_20190828173350}\medskip
\end{minipage}
\begin{minipage}[b]{.48\linewidth}
  \centering
  \centerline{\includegraphics[width=4.4cm]{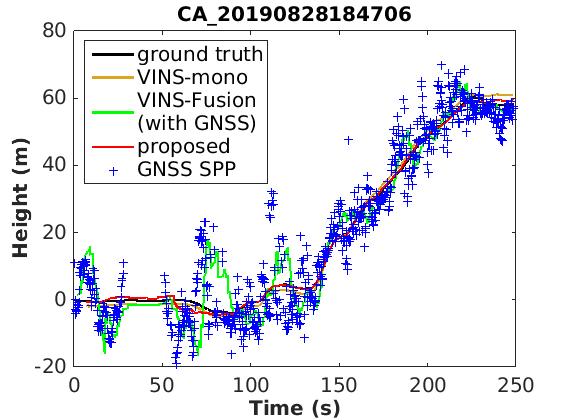}}
  \centerline{(c) CA\_20190828184706}\medskip
\end{minipage}
\hfill
\begin{minipage}[b]{0.48\linewidth}
  \centering
  \centerline{\includegraphics[width=4.4cm]{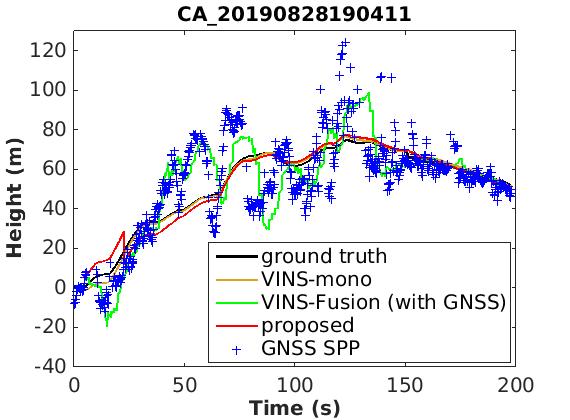}}
  \centerline{(d) CA\_20190828190411}\medskip
\end{minipage}
\caption{Comparison of trajectories in height direction. In figure (a) and (b) a few very noisy GNSS SPP results that lie outside the figure are omitted. All trajectories are aligned with the ground truth.}
\label{fig_height}
\end{figure}
\subsection{Evaluation on Accuracy of Trajectories}

Our proposed approach is compared with state-of-the-art VI-SLAM approach VINS-mono \cite{qin2018vins}, GNSS single point positioning results computed using RTKLIB \cite{takasu2011rtklib}, and the loose coupling approach VINS-Fusion (with GNSS) \cite{qin2019b}. The accuracies of the above approaches are compared in Table \ref{table_average_error}. In Table \ref{table_average_error}, all the four approaches are evaluated by ourselves. Although the evaluation results of VINS-mono \cite{qin2018vins} is presented  in \cite{wen2020urbanloco}, it is reevaluated by us because the incorrect part of ground truth is excluded in the sequence CA\_20190828190411. \emph{GNSS SPP} is the single point positioning result produced using RTKLIB \cite{takasu2011rtklib}. For VINS-Fusion (with GNSS) \cite{qin2019b}, the inputted GNSS positions are the single point positioning results from RTKLIB \cite{takasu2011rtklib}, i.e. the results of \emph{GNSS SPP}. In this subsection, for our proposed approach, the second method to remove the noisy GNSS measurements described in Sect. \ref{subsec_better} is applied. For fairness of comparison, we employ the frontal camera and the IMU in VINS-mono \cite{qin2018vins}, VINS-Fusion (with GNSS) \cite{qin2019b} and our proposed approach. Also for fairness of comparison, because the resulting trajectory of VINS-mono \cite{qin2018vins} is aligned to the ground truth by a rigid body transformation \cite{sturm2012benchmark} before evaluation, the other three approaches are also aligned to the ground truth by a rigid body transformation before evaluation. For each approach the loop closure option is turned off. For \emph{GNSS SPP} and our proposed approach, the same raw GNSS measurements are utilized, including the measurements from both GPS and GLONASS. Same as the evaluation of VINS-mono \cite{qin2018vins} in paper \cite{wen2020urbanloco}, our proposed approach is also evaluated at about half-real time playback rate, and VINS-mono \cite{qin2018vins} is reevaluated at half-real time playback rate. The comparisons of trajectories among different approaches w.r.t the east and north directions, as well as w.r.t. the height direction are also shown in Fig.\ref{fig_tra} and Fig.\ref{fig_height} respectively. 

From Table \ref{table_average_error}, Fig.\ref{fig_tra} and Fig.\ref{fig_height} we can see that our proposed approach outperforms the other three approaches in terms of both translation and rotation accuracy. Table \ref{table_average_error} reports the mean absolute error (MAE) in translation and in rotation, the root mean square error (RMSE) in translation, as well as the trajectory completeness. RMSE in translation is precisely the absolute trajectory error (ATE) presented in \cite{sturm2012benchmark}. Trajectory completeness is computed as follows: We sample the time interval every 0.1 second from the beginning to the end of each sequence. For every sampled moment, if and only if position is successfully calculated at any moment that is within 3 seconds before or after it, the moment is regarded as successfully positioned. Trajectory completeness is the ratio of the number of successfully positioned sampled moments to the number of all the sampled moments. 
From Table \ref{table_average_error} the most evident conclusion that can be drawn is that the translation RMSE of our proposed approach is dramatically smaller than those of other approaches. 

\begin{figure}[htb]
\begin{minipage}[b]{.48\linewidth}
  \centering
  \centerline{\includegraphics[width=4.1cm]{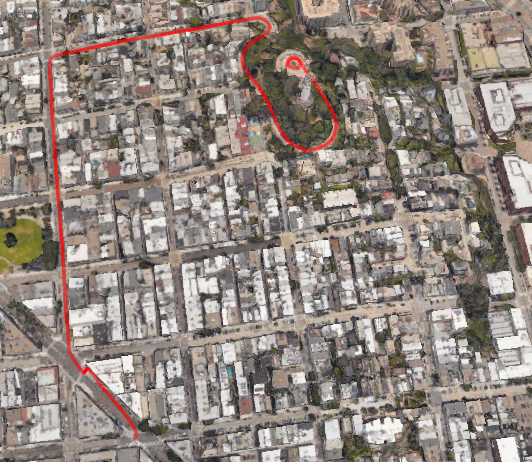}}
  \centerline{(a) CA\_20190828184706}\medskip
\end{minipage}
\hfill
\begin{minipage}[b]{0.48\linewidth}
  \centering
  \centerline{\includegraphics[width=4.1cm]{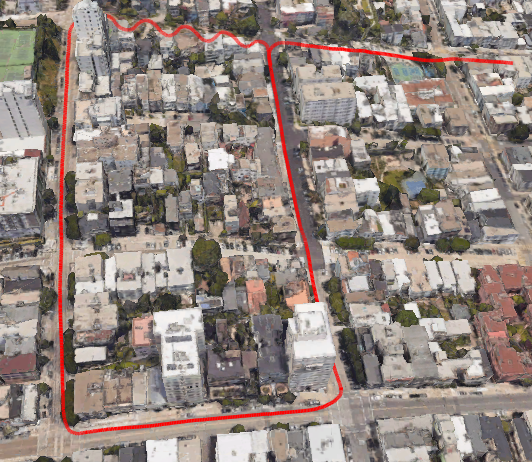}}
  \centerline{(b) CA\_20190828190411}\medskip
\end{minipage}
\caption{Our resulting trajectories of (a) CA\_20190828184706 and (b) CA\_20190828190411 projected onto Google Earth.}
\label{fig_google}
\end{figure}

Through tightly coupling raw GNSS measurements, our approach achieves global positioning results, whose resulting trajectories on the two sequences are projected onto Google Earth as shown in Fig.\ref{fig_google}.

\subsection{Comparison on Two Noisy GNSS Measurements Removal Methods}
\label{ssec:discussion}
A comparison on the two noisy GNSS measurements removal methods described in Sect. \ref{subsec_better} is supported here in terms of absolute trajectory error (ATE) and time consumption. From Table \ref{table_options} we can see that the second one has a better accuracy, at the cost of more time consumption. 
\begin{table}[htb]
\caption{Comparison of ATE and mean time consumption (MTC)}
\label{table_options}
\centering
\begin{threeparttable}
\begin{tabular}{*{5}{c}}
\bottomrule
Sequence&Length&Method&ATE (m)&MTC (ms)\\
\midrule
CA\_20190828-&\multirow{2}{*}{5.9km}&GNSS solver&19.081&\textbf{0.389}\\
155828&&Mixed solver&\textbf{14.328}&24.663\\
\midrule
CA\_20190828-&\multirow{2}{*}{3.2km}&GNSS solver&20.280&\textbf{0.341}\\
173350&&Mixed solver&\textbf{14.337}&26.080\\
\midrule
CA\_20190828-&\multirow{2}{*}{1.8km}&GNSS solver&\textbf{3.395}&\textbf{0.336}\\
184706&&Mixed solver&3.590&30.141\\
\midrule
CA\_20190828-&\multirow{2}{*}{1.0km}&GNSS solver&5.941&\textbf{0.347}\\
190411&&Mixed solver&\textbf{4.455}&29.473\\
\bottomrule
\end{tabular}
\begin{tablenotes}
        \fontsize{8pt}{10pt}
		\item Here \emph{mean time consumption (MTC)} means the mean time consumption of one execution of excluding the noisy GNSS measurements. \emph{GNSS solver} and \emph{Mixed solver} represent the first method and the second method in Sect. \ref{subsec_better} respectively.
\end{tablenotes}
\end{threeparttable}
\end{table}

\section{CONCLUSIONS}

In this paper, we propose an optimziation-based visual-inertial SLAM tightly coupled with raw GNSS measurements. Feature points, IMU measurements as well as pseudoranges and Doppler shift measurements from GNSS, which are captured at different moments, are integrated by means of optimization in a sliding window. Experimental results prove that our proposed approach outperforms state-of-the-art visual-inertial SLAM, GNSS single point positioning, as well as the loose coupling approach \cite{qin2019b} on public dataset. 






\end{document}